%% file: Self_Spec.tex
\documentclass[english]{article}
\usepackage[T1]{fontenc}
\usepackage{geometry}
\usepackage[unicode=true,
 bookmarks=false,
 breaklinks=false,
 pdfborder={0 0 1},
 colorlinks=false]{hyperref}
\hypersetup{colorlinks,
linkcolor=red,
anchorcolor=blue,
citecolor=blue,
urlcolor=blue}
\usepackage{url}
\usepackage{amsmath,amsfonts,bm,amssymb,amsthm}
\usepackage{natbib}
\usepackage{algorithm}
\usepackage{algpseudocode}
\usepackage{booktabs}
\usepackage{multirow}
\usepackage{graphicx}
\usepackage{xcolor}
\usepackage{booktabs}
\usepackage{makecell}
\usepackage{dsfont}
\usepackage{fontawesome5}

\definecolor{yx}{RGB}{0,0,200}
\definecolor{cc}{RGB}{231,117,0}

\allowdisplaybreaks

\newcommand{\cT}{\mathcal{T}}

\newcommand{\cV}{\mathcal{V}}

\newcommand{\KL}{\mathsf{KL}}

\newcommand{\TV}{\mathsf{TV}}

\newcommand{\MAL}{\mathsf{MAL}}

\newcommand{\sg}{\operatorname{sg}}

\newcommand{\E}{\mathbb{E}}

\newcommand{\target}{}
\newcommand{\eos}{\texttt{<EOS>}}

\newcommand{\pos}[1]{\left[#1\right]_{+}}

\title{Training Parallel Speculative Draft Models by Directly Minimizing Expected Decoding Rounds}

\author{
  Yunxiao Zhao\thanks{Department of Industrial and Operations Engineering, University of Michigan, Ann Arbor, MI, USA.
  {Email: \texttt{\{zhyx,cxcai\}@umich.edu}.}
  }
  \and
  Changxiao Cai\footnotemark[1]
}
\date{\today}

\begin{document}

\theoremstyle{plain} 
\newtheorem{lemma}{\bf Lemma} 
\newtheorem{proposition}{\bf Proposition}
\newtheorem{theorem}{\bf Theorem}
\newtheorem{corollary}{\bf Corollary} 
\newtheorem{claim}{\bf Claim}

\theoremstyle{remark}
\newtheorem{assumption}{\bf Assumption} 
\newtheorem{definition}{\bf Definition} 
\newtheorem{condition}{\bf Condition}
\newtheorem{property}{\bf Property} 
\newtheorem{example}{\bf Example}
\newtheorem{fact}{\bf Fact}
\newtheorem{remark}{\bf Remark}

\maketitle 
\setcounter{tocdepth}{2}

\input{abstract}

\begin{center}
  \faGithub\ \url{https://github.com/y-x-zhao/AngelSpec-EDR}
\end{center}


\tableofcontents{}

\input{intro}
\input{notation}

\input{result}
\input{related}
\input{loss}
\input{experiments}
\input{conclusion}

\section*{Acknowledgements}
C.\ Cai is supported by the NSF CAREER Award CCF-2541600 and the NSF grant DMS-2515333.

\bibliographystyle{apalike}
\bibliography{ref}

\appendix
\input{algorithm}
\input{proof}
\input{appendix}
\input{exp_details}
\end{document}

%% file: abstract.tex
\begin{abstract}
Speculative decoding accelerates large language model inference by using a low-cost draft model to propose tokens that the full-size target model verifies in parallel. 
Parallel and semi-autoregressive (semi-AR) drafters improve drafting efficiency by proposing an entire block in a single forward pass, but training them raises a new difficulty: the draft distribution for a given position depends on where the decoding round starts, and where rounds start depends on how many tokens earlier rounds accepted.
Existing training objectives typically rely on block-local surrogates that ignore this cross-round coupling, and therefore do not directly optimize the global decoding efficiency.
In this work, we develop a theoretical framework for training and evaluating these drafters by representing speculative decoding as a Markov reward process.
This formulation yields the Expected Decoding Rounds (EDR) objective, which weights local rejection costs by state occupancies and exactly equals the expected number of decoding rounds.
Unlike prior surrogate objectives, EDR introduces no auxiliary hyperparameters.
We then derive an exact temporal-difference gradient that supports unbiased stochastic optimization from target-model rollouts.
The same framework also yields an exact offline evaluator for round counts, enabling paired drafter comparisons on shared target rollouts without running speculative decoding.
Finetuning two state-of-the-art drafters, DSpark and DFly, with EDR consistently improves mean accepted length and outperforms existing training objectives across nine benchmarks spanning math reasoning, code generation, and chat.

\end{abstract}

%% file: intro.tex
\section{Introduction}
\label{sec:intro}

Autoregressive (AR) generation is a central bottleneck in large language model (LLM) inference.
Since each token can be produced only after its preceding tokens are available, the number of function evaluations required to generate a text sequence grows linearly with its length. 

Speculative decoding~\citep{leviathan2023fast,chen2023accelerating} reduces this sequential cost by using a lightweight \textit{draft model} to propose multiple candidate tokens, which the full-size \textit{target model} verifies in parallel in a single forward pass.
With rejection-sampling verification, the generated sequence follows the target model's distribution exactly. 
Therefore, speculative decoding accelerates inference without sacrificing quality, and the goal is to reduce expensive target model verifications---equivalently, decoding rounds---while keeping drafting itself inexpensive.
Early work employed AR drafters~\citep{li2024eagle1} that predict draft tokens sequentially, so the latency contributed in the drafting phase grows linearly with the draft block size.
To reduce the sequential work of drafting, parallel drafters~\citep{cai2024medusa,dflash} have emerged to propose multiple draft tokens in one forward pass, typically built on masked diffusion language models~\citep{sahoo2024simple,shi2024simplified}.
Their semi-AR variants~\citep{dspark, angelspec} add limited sequential dependence within each draft block to improve the draft quality.
Parallel drafting supports longer draft blocks and deeper drafter networks, with potential to significantly accelerate the inference of a large target model.

\paragraph{Challenges in parallel drafter training: cross-round dependence.} 
For a fixed architecture, a drafter's quality affects inference speed primarily through the number of decoding rounds, so the natural training objective is to minimize the expected number of rounds.

For an AR drafter, the proposal at position $t$ is conditioned on the full realized prefix $x_{<t}$ regardless of where the current round started, so its acceptance probability depends only on this prefix.
As a result, the standard local training objective for AR drafters---matching the target model's next-token distribution at every prefix---is directly aligned with this goal \citep{yin2024theoretical}.
In contrast, for a parallel or semi-AR drafter, the proposal can instead depend on an earlier prefix $x_{\leq n}$ with $n < t$ (in a round starting at $n$), with access to only some or none of the intermediate tokens.
Consequently, the same output position can have different draft distributions depending on where the round starts.

This additional dependence breaks the direct link between local distribution matching and global decoding efficiency. 
The draft for position $t$ from a round starting at $n$ matters only if the decoding trajectory first reaches a round starting at $n$ and then accepts all preceding draft tokens in that round up to position $t-1$.
Both events depend on earlier acceptance outcomes induced by the drafter itself.
Thus, changing the drafter affects both acceptance within a round and which draft distributions are encountered in subsequent rounds, making the training objective inherently coupled across rounds. 

However, existing objectives for parallel and semi-AR drafters do not account for these cross-round dependencies and instead rely on local surrogates.
Examples include the expected number of tokens accepted within a single round \citep{e2e} and draft-target distributional discrepancies under hand-crafted weights \citep{dflash,dpace}. 
%
%
This motivates the central question:
\begin{center}
\emph{Can we train parallel and semi-AR drafters by directly minimizing the expected number of decoding rounds?}
\end{center} 

\paragraph{Our contributions.} 
We answer this question affirmatively.
Our approach relies on two key observations. First, conditioned on the output sequence, speculative decoding with a parallel or semi-AR drafter can be formulated as a Markov reward process (MRP) \citep{howard1971dynamic}: each state records where the current round started and which position is being verified, and each rejection costs one round. Second, since rejection-sampling verification ensures that the output follows the same target distribution for every drafter, averaging over target rollouts recovers the expected number of rounds of actual decoding.
Together, these insights allow us to capture the dependencies across decoding rounds explicitly and characterize the expected number of decoding rounds exactly, yielding a unified framework for training and evaluating drafters.

Building on this framework, we introduce the \textit{Expected Decoding Rounds} (EDR) objective, whose minimization is equivalent to minimizing the expected number of decoding rounds.
We then derive a temporal-difference gradient estimator, enabling drafter training from target-model rollouts. 
EDR applies across drafter architectures and introduces no hyperparameters.
We further prove that drafters maximizing the expected accepted length for each round can be strictly suboptimal for EDR, theoretically justifying the advantages of our global objective.
The framework also yields an exact offline evaluator of expected round counts, enabling comparisons of drafters on the same target-model rollouts without running online speculative decoding.

Empirically, we finetune two state-of-the-art (SOTA) drafters, DSpark \citep{dspark} and DFly \citep{angelspec}, with EDR
while keeping their architectures and inference procedures unchanged. Across nine benchmarks spanning mathematical reasoning, code generation, and chat, a single epoch of EDR finetuning consistently improves mean accepted length over the original drafters and existing training objectives.

Our main contributions are summarized as follows:
\begin{itemize}
    \item \textbf{A Markov reward process formulation.} We provide a theoretical framework that exactly characterizes expected speculative decoding rounds for parallel and semi-AR drafters.
\item \textbf{An exact and trainable global objective.}
    We introduce EDR, establish its equivalence to minimizing the expected number of decoding rounds, derive a tractable gradient for drafter training, and prove its advantage over local surrogates.
    %

    \item \textbf{An exact offline evaluator.}
    We provide an offline evaluator of the expected number of decoding rounds, enabling paired evaluation of drafters on the same target-model rollouts.

    \item \textbf{Consistent empirical gains.}
    EDR finetuning improves two SOTA drafters across nine benchmarks, outperforming existing training objectives.
\end{itemize}

%% file: notation.tex
\section{Preliminaries}
\label{sec:notation}

Let $\cV$ be a vocabulary of ordinary tokens and
$\overline{\cV}:=\cV\cup\{\eos\}$, where $\eos$ denotes the end-of-sequence token.
For two distributions $p,q$, define the total variation (TV) distance
$ \TV(p,q) :=\sum_{v\in\overline{\cV}}\pos{p(v)-q(v)} $, where $[z]_+ := \max(0, z)$.
For any sequence $x=(x_1, x_2, \ldots)$, we denote $x_{\leq a} = (x_1, x_2, \ldots, x_a)$, $x_{<a} := x_{1:a-1}$, and $x_{a:b} := (x_a, x_{a+1}, \ldots, x_b)$ for $1\leq a\le b$.
For $ a > b $, $ x_{a:b} := \emptyset $.

\paragraph{Target distribution.}
Let $p$ denote the distribution of the target model.
It generates tokens autoregressively as
$x_t\sim p_t(\cdot\mid x_{<t})$ for $t \geq 1$ and stops upon sampling $\eos$. 
Denote by $L:=\inf\{\ell\ge0:x_{\ell+1}=\eos\}$ the sequence length excluding $\eos$.
We assume $ L $ has a universal constant upper bound for $x\sim p$.
A complete sequence $x=(x_{1:L},\eos)$ has probability $\prod_{t=1}^{L} p_t(x_t\mid x_{<t}) p_{L+1}(\eos\mid x_{1:L}),$ and we call $x$ \emph{target-supported} if $p(x)>0$.

\paragraph{Speculative decoding.}

Speculative decoding generates a token sequence $x$ in rounds.
Consider a round that starts from a committed prefix $x_{1:n}$.
A drafter $q$ first proposes a block of $B$ \textit{draft tokens} $\widetilde x_{n+1:n+B}$ with $\widetilde x_t \sim q_t$.
For simplicity, we write $q_t$ for the draft distribution at position $t$ in the current round, suppressing its dependence on the round start $n$, committed prefix $x_{1:n}$, and preceding within-block proposals $\widetilde x_{n+1:t-1}$.
We will make this explicit in the next paragraph.

The target model then verifies the draft block from left to right in a single forward pass.
When position $t$ is verified, all earlier drafts in the block have been accepted, so the current prefix is $x_{<t}=(x_{1:n},\widetilde x_{n+1:t-1})$. Write $p_t = p_t(\cdot\mid x_{<t})$. The draft $\widetilde x_t$ is accepted with probability $\min\{ 1,{p_t(\widetilde x_t)}/{q_t(\widetilde x_t)} \}$, in which case it is committed as $x_t = \widetilde x_t$. 
Otherwise, it is replaced with a \textit{correction token} $x_t\sim r_t$ drawn from the residual distribution $r_t(v) := [p_t(v) - q_t(v)]_+/\TV(p_t,q_t)$, $v\in\overline{\cV}$. The remaining drafts in the block are discarded, and the next round begins from the updated prefix $x_{1:t}$.
%
If all $B$ draft tokens are accepted, the same forward pass also yields one additional \emph{bonus token} $x_{n+B+1}\sim p_{n+B+1}(\cdot\mid x_{\leq n+B})$, and the next round starts after it.
Generation stops when $\eos$ is committed.
This rejection-sampling verification ensures that the output sequence satisfies $x\sim p$ for any drafter $q$ \citep{leviathan2023fast}.

\paragraph{Draft distribution.}

Drafters differ in what $q_t$ conditions on.
An AR drafter proposes each token from the full prefix, whereas parallel and semi-AR drafters propose the whole block from the prefix committed at the start of the round.
We now refine the shorthand $q_t$ to make this explicit.
For a round starting from a committed prefix $x_{1:n}$, let $q_{n,t}(\cdot\mid x_{1:n};\widetilde x_{n+1:t-1})$ denote the draft distribution for position $t=n+1,\dots,n+B$.
The subscript $t$ identifies the position being predicted, and $n$ records the position where the round starts.
The semicolon separates the committed prefix $x_{1:n}$ from the preceding within-block proposals $\widetilde x_{n+1:t-1}$, which the drafter may use in whole or in part.

This notation covers AR, parallel, and semi-AR drafters.
An AR drafter is the special case $q_{n,t}(\cdot\mid x_{1:n};\widetilde x_{n+1:t-1})=q_t(\cdot\mid x_{1:n},\widetilde x_{n+1:t-1})$, which conditions on the concatenated prefix and has no further dependence on the round start $n$.
A parallel drafter samples the block independently given the committed prefix $x_{1:n}$, i.e., $q_{n,t}(\cdot\mid x_{1:n};\widetilde x_{n+1:t-1})=q_{n,t}(\cdot\mid x_{1:n})$.
A semi-AR drafter allows partial dependence on preceding within-block drafts while maintaining parallel prediction across some positions.
For parallel and semi-AR drafters, the same current prefix can therefore induce different draft distributions depending on the round start $n$. This dependence is the main difference from AR drafters and the central challenge this work addresses.
We give the Pseudocode for speculative decoding process we considered in Algorithm~\ref{alg:self-sd} in Appendix~\ref{sec:algorithm}.

\paragraph{Decoding rounds and mean accepted length.}
Each decoding round requires one forward pass of the target model.
Let $\cT^q$ denote the number of decoding rounds needed to produce a complete output under drafter $q$, including the one that produces $\eos$.
A common metric is the mean accepted length (MAL), the ratio of the expected output length to the expected number of rounds,
\begin{equation}
    \MAL(q)
    :=
    \frac{\E[L + 1]}{\E[\cT^q]},
    \label{eq:mal-definition}
\end{equation}
where $L+1$ is the length of the complete output counting all committed tokens, including correction tokens, bonus tokens, and the terminal $ \eos $ token.
As the output sequence follows the target distribution, $\E[L]$ is independent of the drafter $q$ for a fixed target model and evaluation corpus. Consequently, maximizing $\MAL(q)$ is equivalent to minimizing $\E[\cT^q]$.


%% file: result.tex
\section{Main Results}
\label{sec:EDR}

This section presents our main results. 
Section~\ref{sec:pathwise-transition} presents a Markov reward process (MRP) formulation for speculative decoding.
Section~\ref{sec:weighted-cost-edr} introduces the EDR objective and Section~\ref{sec:EDR-grad} presents a temporal-difference form of its gradient.
Section~\ref{sec:offline-evaluation} gives offline estimators for the decoding statistics, and suboptimality of block-local objectives is discussed in Section~\ref{sec:separation}.

\subsection{Speculative Decoding as a Markov Reward Process}
\label{sec:pathwise-transition}



Recall that $\cT^q$ denotes the number of decoding rounds in a speculative decoding run with drafter $q$, and let $X=(X_1,\dots,X_L,\eos)$ be the output sequence of the same run. Since speculative decoding ensures $X\sim p$ for any drafter $q$, the overall expected number of decoding rounds satisfies
\begin{align}\label{eq:expected-rounds-conditional}
 \E[\cT^q]
    =
    \E_{X\sim p}[\tau^q(X)],
    \qquad \text{ with }
    \tau^q(x):=\E[\cT^q\mid X=x],
\end{align}
where $\tau^q(x)$ is the expected number of rounds conditioned on producing the sequence $x$.
Therefore, it suffices to characterize $\tau^q(x)$ for any target-supported sequence $x$.

\paragraph{Markov reward process.}
In light of this, we fix a target-supported sequence $x=(x_1,\dots,x_L,\eos)$.
Conditioned on the output $x$ of a speculative decoding run, the remaining randomness is which positions are filled by accepted drafts and which by the target model after a rejection, and this determines the number of decoding rounds $\cT^q(x)$.

This can be described as a Markov chain on states $(n,t)$ for $0\le n<t\le L+1$. (see Appendix~\ref{sec:markov-chain} for a detailed derivation). 
State $(n,t)$ indicates that the current round started after $n$ (i.e., $x_{1:n}$ was committed) and is verifying the draft for position $t$.
Decoding starts at $(0,1)$ and terminates when $\eos$ is produced at position $L+1$.

Note that at state $(n,t)$ with $t\leq n+B$, the draft for position $t$ is drawn from $q_{n,t}(\cdot \mid x_{1:n};x_{n+1:t-1})$. By Bayes' rule, given that position $t$ outputs $x_t$, the probablity that it was produced by an accepted draft is
\begin{equation}
    A_{n,t}^{q}(x_{\leq t})
    :=
    \min\bigg\{
        1,
        \frac{
            q_{n,t}(
                x_t\mid x_{1:n};x_{n+1:t-1}
            )
        }{
            p_t(x_t\mid x_{<t})
        }
    \bigg\}.
    \label{eq:path-accept-prob}
\end{equation}
Denote by $R_{n,t}^{q}:=1-A_{n,t}^{q}$ the rejection probability. 
%
Acceptance continues the current round, whereas rejection starts a new round after position $t$, provided that $x_t \neq \eos$. Therefore, the Markov chain has two transitions:
\begin{equation}
    (n,t)
    \longrightarrow
    \begin{cases}
        (n,t+1), & \text{with probability}~~A_{n,t}^{q}(x_{\leq t}),\\
        (t,t+1), & \text{with probability}~~R_{n,t}^{q}(x_{\leq t}).
    \end{cases}
    \label{eq:mrp-transition}
\end{equation}
We set $A_{n,t}^{q}=0$ and $R_{n,t}^{q}=1$ for $t>n+B$, so the same transition rule covers the case when all tokens in a draft block are accepted
and a new round starts after that.

Let $\omega_{n,t}^{q}(x)$ denote the \textit{occupancy} of state $(n,t)$: the probability that this state is visited, conditioned on the output sequence being $x$.
It is not hard to see that $\omega_{n,t}^{q}(x)$ depends on $x$ only through $x_{<t}$ and hence we write $\omega_{n,t}^{q}(x_{<t})=\omega_{n,t}^{q}(x)$.
Starting from $\omega_{0,1}^{q}(x)=1$, the occupancies satisfy the forward recursion:
\begin{subequations}
\label{eq:occupancy-recursion}
\begin{align}
    \omega_{n,t+1}^{q}(x_{\leq t})
    &=\omega_{n,t}^{q}(x_{<t})A_{n,t}^{q}(x_{\leq t}),
    \quad n<t \leq L,
    \label{eq:occupancy-continue}\\
    \omega_{t,t+1}^{q}(x_{\leq t})
    &=\sum_{n=0}^{t-1}\omega_{n,t}^{q}(x_{<t})R_{n,t}^{q}(x_{\leq t}),
    \quad 1\le t\le L.
    \label{eq:occupancy-restart}
\end{align}
\end{subequations}

Since every round begins at a diagonal state $(t,t+1)$, the number of rounds equals the number of diagonal states visited, yielding the following exact expression (see Appendix~\ref{sec:proof-pathwise-rounds} for proof).
\begin{proposition}
\label{prop:pathwise-rounds}
For any target-supported sequence $x=(x_{1:L},\eos)$ and any drafter $q$, the expected number of decoding rounds to produce $x$ is
\begin{equation}
    \tau^q(x)
    = \sum_{t=0}^{L}\omega_{t,t+1}^{q}(x_{\leq t}) 
    = 
    1+
    \sum_{t=1}^{L}\sum_{n=0}^{t-1}
    \omega_{n,t}^{q}(x_{<t})R_{n,t}^{q}(x_{\leq t}).
    \label{eq:pathwise-weighted-transition}
\end{equation}
\end{proposition}

The second identity  has a direct MRP interpretation. Rejection starts a new round and incurs unit cost, while acceptance incurs zero cost. Hence, the rejection probability $R_{n,t}^{q}$ is the expected cost of visiting state $(n,t)$, and the occupancy-weighted sum of costs gives the round count.

\subsection{The EDR Objective}
\label{sec:weighted-cost-edr}

By \eqref{eq:expected-rounds-conditional}, averaging~\eqref{eq:pathwise-weighted-transition} over $x\sim p$ provides an exact expression for the overall expected number of decoding rounds $\E[\cT^q]$. However, each rejection probability $R_{n,t}^{q}(x_{\le t})$ uses the target only through the single realized token $x_t$, even though the full distribution $p_t(\cdot\mid x_{<t})$ is available.
Hence, we average over this token under the target distribution $p_t$, conditioned on $x_t$ being a non-EOS token. 
\begin{definition}
Define the \emph{local rejection cost} of state $(n,t)$, for $0\le n<t\le L$, as
    \begin{align}\label{eq:weighted-cost}
    c_{n,t}^{q}(x_{<t}) := \mathop{\E}_{X_t \sim p^\prime_t(\cdot \mid x_{<t})}[
        R_{n,t}^{q}(x_{<t}, X_t)
    ]
    \end{align}
where $p^\prime_t(v\mid x_{<t}) := p_t(v\mid x_{<t})/(1-p_t(\eos\mid x_{<t}))$ for $v\neq \eos$ is the target distribution normalized over non-EOS tokens. For $t \leq n+B$, its value is given by
\begin{align}
c_{n,t}^{q}(x_{<t})=
    \frac{1
    }{1-p_t(\eos\mid x_{<t})
    }\sum_{v\in\cV}
            [p_t(v\mid x_{<t})
            -q_{n,t}(
                v\mid x_{1:n};x_{n+1:t-1}
            )]_{+},
\end{align}
and for $t>n+B$, it is equal to $1$.
Note that $c_{n,t}^{q}$ is well-defined because every position $t\le L$ outputs a non-EOS token so $p_t(\eos\mid x_{<t})<1$.
We denote $ c_{n, L+1}^q(x) = 0 $ for notational convenience, since the EOS token incurs no cost.
\end{definition}
%
%

In words, the cost $c_{n,t}^{q}(x_{<t})$ is the rejection probability associated with state $(n,t)$, given that the committed output token $ x_t $ is non-EOS.
It measures the local distributional discrepancy between the drafter $q_{n,t}$ and target $p_t$ for position $t$, given by the TV distance between the draft and target distributions for position $t$, excluding EOS and normalized by the target's non-EOS mass.
Excluding EOS reflects the fact that when the target model produces EOS, it never starts
another decoding round and incurs no penalty for draft rejection.
%
%
%

We now introduce the \emph{Expected Decoding Rounds (EDR)} objective,
which exactly equals the overall expected number of decoding rounds, as formalized in Theorem~\ref{thm:exact-edr} below.
The proof can be found in Appendix~\ref{sec:proof-exact-edr}.  
\begin{theorem}
\label{thm:exact-edr}
For any drafter $q$, the EDR objective equals the expected number of decoding rounds:
\begin{equation}
    \mathcal L_{\mathrm{EDR}}(q)
    :=
    1+
    \E_{X\sim p}\!\left[
        \sum_{t=1}^{L}\sum_{n=0}^{t-1}
        \omega_{n,t}^{q}(X_{<t})c_{n,t}^{q}(X_{<t})
    \right] = \E_{X\sim p}[\tau^q(X)] = \E[\cT^q].
    \label{eq:exact-decoding-rounds}
\end{equation}
\end{theorem}
The EDR objective is a sum of local rejection costs $c_{n,t}^{q}$, measuring the local distributional discrepancy between the drafter $q_{n,t}$ and target $p_t$, each weighted by the corresponding occupancy $\omega_{n,t}^{q}$.
The quantity inside the expectation, which we call the \emph{per-sequence EDR loss},
\begin{align}
    \widehat{\mathcal L}_{\mathrm{EDR}}(q;x)
    &:=
    1+
    \sum_{t=1}^{L}\sum_{n=0}^{t-1}
    \omega^q_{n,t}(x_{<t})\,c^q_{n,t}(x_{<t})
    \label{eq:edr-training-estimator}
\end{align}
can be computed exactly given a target-sampled sequence $x$ via the forward occupancy recursion \eqref{eq:occupancy-recursion} and the local rejection cost \eqref{eq:weighted-cost}.
Therefore, averaging over a batch of target rollouts provides an empirical EDR loss for drafters that directly targets minimizing the expected number of rounds.

We make several important remarks.
\begin{itemize}
    \item 
\emph{Replacement of $R_{n,t}^{q}$ with $c_{n,t}^{q}$.}
The EDR loss \eqref{eq:edr-training-estimator} replaces $R_{n,t}^{q}(x_{\leq t} )$ in \eqref{eq:pathwise-weighted-transition} with its conditional expectation $c_{n,t}^{q}(x_{<t} )$. 
For an individual rollout, it generally differs from $\tau^q(x)$, but they agree after averaging over $x\sim p$.
In return, the EDR loss uses the next-token distribution rather than one sampled token, so its gradient aggregates information across the vocabulary.

\item \emph{Exact occupancy weighting.} Prior surrogate objectives \citep{dflash,dpace} do not account for EOS termination in their local discrepancy terms, and more importantly, weight them in a heuristic way. 
These weights can be interpreted as fixed surrogates for the drafter-dependent occupancies $\omega_{n,t}^{q}$. 
In contrast, EDR combines EOS-aware local costs with the actual occupancies, yielding an exact objective for the expected number of decoding rounds.

\item \emph{Parallel vs.~AR drafters.} For an AR drafter, $q_{n,t}(\cdot\mid x_{1:n};x_{n+1:t-1})=q_t(\cdot\mid x_{<t})$ for all $n<t$. Without a block cap ($B=\infty$), EDR reduces to
\begin{align}
\mathcal L_{\mathrm{EDR}}(q)
    =
    1+
    \E_{X\sim p}\bigg[
        \sum_{t=1}^{L}c_t(X_{<t})
    \bigg] \quad \text{with} \quad c_t(x_{<t}):=\frac{
        \sum_{v\in\cV}
            [p_t(v\mid x_{<t})
            -q_t(
                v\mid x_{<t}
            )]_{+}
    }{
        1-p_t(\eos\mid x_{<t})
    }.
    \label{eq:ar-edr}
\end{align}
EDR hence decouples into local prefix costs, each minimized by matching $q_t$ to $p_t$.
This recovers the result of \citet[Theorem~1]{yin2024theoretical} established for AR drafters and extends it to sequences that terminate with EOS.
A finite block size adds only a mild dependence on $n$ through the boundary
states.
Therefore, Theorem~\ref{thm:exact-edr} highlights the key difference between parallel and AR drafters, and EDR captures the cross-round dependencies exactly.
\end{itemize}

\subsection{TD-Form Gradient of EDR}
\label{sec:EDR-grad}

%
The per-sequence EDR loss in \eqref{eq:edr-training-estimator} depends on the drafter $q$ through both the local costs
$c_{n,t}^{q}$ and the occupancies $\omega_{n,t}^{q}$.
Differentiating it with respect to $q$ directly is expensive because it must backpropagate through the entire forward recursion for occupancy $\omega_{n,t}^q$~\eqref{eq:occupancy-recursion}.
Fortunately, the MRP formulation gives the gradient in a temporal-difference (TD) \citep{sutton1988learning} form that uses only local derivatives.
%
%

For a fixed target-supported $x$, consider the MRP with transitions~\eqref{eq:mrp-transition} and costs $c_{n,t}^{q}(x_{<t} )$ from \eqref{eq:weighted-cost}.
Define the value function $V_{n,t}^{q}(x)$, the expected cumulative cost starting from state $(n,t)$. 
The value function satisfies the Bellman equation \citep{bellman1966dynamic}
\begin{align}
    V_{n,t}^{q}(x)
    &=
    c_{n,t}^{q}(x_{<t})
    +A_{n,t}^{q}(x_{\leq t})V_{n,t+1}^{q}(x)
    +R_{n,t}^{q}(x_{\leq t})V_{t,t+1}^{q}(x),
    \qquad 0\leq n < t\le L,
    \label{eq:V-bellman}
\end{align}
with the terminal condition $V_{n,L+1}^{q}(x)=0$ for all $0\leq n\le L$. 
Unlike the cost and occupancy, the value function $V_{n,t}^{q}(x)$ depends on the whole sequence $x$.
The value function of the initial state gives the per-sequence EDR loss:
\begin{equation}
    V_{0,1}^{q}(x)
    =
    \widehat{\mathcal L}_{\mathrm{EDR}}(q;x)-1.
    \label{eq:weighted-cost-from-V}
\end{equation}

\begin{algorithm}[t]
\caption{One EDR training step ($\sg(\cdot)$ denotes the stop-gradient operator)}
\label{alg:edr}
\begin{algorithmic}[1]
\Require drafter $q_\theta$, target model $p$, block size $B$
\State For each prompt $i=1,\dots,M$, sample an independent target rollout
  $x^{(i)}=(x^{(i)}_{1:L_i},\eos)$ from the target model $p$
\State Compute $p_t(\cdot\mid x^{(i)}_{<t})$ for all $i$ and $t$ in one batched target forward pass
\For{$i=1,\dots,M$ (in parallel)}
    \State Compute $A^{(i)}_{n,t}$ and $c^{(i)}_{n,t}$ from $q_\theta$ at all states $(n,t)$
    \State Compute $\omega^{(i)}_{n,t}$ by~\eqref{eq:occupancy-recursion} and
        $V^{(i)}_{n,t}$ by~\eqref{eq:V-bellman}, without gradients
\EndFor
\State $\ell(\theta)\gets M^{-1}\sum_{i=1}^{M}\sum_{t=1}^{L_i}\sum_{n=0}^{t-1}\sg(\omega^{(i)}_{n,t})
    \bigl[c^{(i)}_{n,t}+A^{(i)}_{n,t}\sg(V^{(i)}_{n,t+1}-V^{(i)}_{t,t+1})\bigr]$
\State Update $\theta$ using $\nabla_\theta\ell(\theta)$
\end{algorithmic}
\end{algorithm}

We now present the gradient of the per-sequence EDR loss in the following theorem. The proofs of \eqref{eq:weighted-cost-from-V} and Theorem \ref{prop:edr-gradient} are given in Appendix~\ref{sec:proof-edr-gradient}. 
\begin{theorem}[EDR gradient]
\label{prop:edr-gradient}
We parametrize the drafter $ q = q_{\theta} $ by neural network parameters $ \theta $. 
For any target-supported sequence $x$, the gradient of the per-sequence EDR loss \eqref{eq:edr-training-estimator} is given by
\begin{equation}
    \nabla_\theta\widehat{\mathcal L}_{\mathrm{EDR}}(q_\theta;x)
    =
    \sum_{t=1}^{L}\sum_{n=0}^{t-1}
    \omega_{n,t}^{q}(x_{<t})\,
    \left[
        \nabla_\theta c_{n,t}^{q}(x_{<t})
        +
        \nabla_\theta A_{n,t}^{q}(x_{\leq t})
        \left(
            V_{n,t+1}^{q}(x)-V_{t,t+1}^{q}(x)
        \right)
    \right].
    \label{eq:edr-gradient}
\end{equation}
As a result, one has
\begin{equation}
    \E_{X\sim p}
    \left[
        \nabla_\theta\widehat{\mathcal L}_{\mathrm{EDR}}(q_\theta;X)
    \right]
    =
    \nabla_\theta\mathcal L_{\mathrm{EDR}}(q_\theta).
    \label{eq:edr-gradient-unbiased}
\end{equation}
\end{theorem}

The gradient has a standard one-step TD form, separated into two contributions: the immediate effect on the local rejection cost, and the downstream effect of changing the acceptance probability. The latter is weighted by the value gap, which measures the difference between the cumulative costs of the two possible subsequent states. The occupancies account for how often each state is reached.
 

\paragraph{Implementation.}

We compute the occupancies $\omega^q_{n,t}$ by the forward recursion \eqref{eq:occupancy-recursion} and the value functions $V^q_{n,t}$ by the backward Bellman recursion \eqref{eq:V-bellman}. 
Since they enter the gradient expression~\eqref{eq:edr-gradient} only as weights, we hold them fixed during differentiation.
Gradients are required only through the local quantities $c_{n,t}^{q}$ and $A_{n,t}^{q}$. 

By~\eqref{eq:edr-gradient-unbiased}, averaging~\eqref{eq:edr-gradient} over a batch of i.i.d.\ target rollouts gives an unbiased stochastic gradient of EDR, which directly optimizes the expected number of decoding rounds.
%
One EDR training step is summarized in Algorithm~\ref{alg:edr}. 


\subsection{Exact Offline Evaluation}
\label{sec:offline-evaluation}

The MRP formulation also yields exact offline evaluators for any drafter, computed from target-model rollouts alone without running speculative decoding.
We first give an unbiased estimator of the expected number of decoding rounds, from which we obtain an estimator of the MAL.
We then extend it to a finer-grained statistic, the expected number of rounds that commit at least $k$ tokens, which yields an estimator of the position-wise acceptance rate.

\subsubsection{Expected Decoding Rounds and MAL}

By Proposition~\ref{prop:pathwise-rounds}, the expected number of decoding rounds conditioned on the output $X=x$ equals the sum of the diagonal occupancies.
For a target-supported sequence $x=(x_{1:L},\eos)$, define
\begin{equation}
    \tau_{\mathrm{off}}^{q}(x)
    :=
    \sum_{t=0}^{L}\omega_{t,t+1}^{q}(x_{\le t}).
    \label{eq:offline-rounds}
\end{equation}
This quantity averages out all randomness of speculative decoding given $X=x$, namely draft sampling and verifier randomness, and is computed exactly by a single pass of the forward occupancy recursion.
Given a dataset of target rollouts
$\mathcal D=\{x^{(i)}=(x_{1:L_i}^{(i)},\eos)\}_{i=1}^{N}$ independently sampled from $x^{(i)}\sim p$,
we estimate the expected number of rounds and the MAL by
\begin{equation}
    \widehat{\cT}_{\mathcal D}(q)
    :=
    \frac1N \sum_{i=1}^{N}\tau_{\mathrm{off}}^{q}(x^{(i)})
    \qquad\text{and}\qquad
    \widehat{\MAL}_{\mathcal D}(q)
    :=
    \frac{N^{-1}\sum_{i=1}^{N}(L_i+1)}{\widehat{\cT}_{\mathcal D}(q)}.
    \label{eq:offline-mal}
\end{equation}
Since $\mathbb E[\cT^q]=\mathbb E_{X\sim p}[\tau^q(X)]$, $\widehat{\cT}_{\mathcal D}(q)$ is an unbiased estimator of $\mathbb E[\cT^q]$, and $\widehat{\MAL}_{\mathcal D}(q)$, a ratio of sample means, is a consistent estimator of $\MAL(q)$.
Moreover, since the rollouts do not depend on the drafter, all drafters can be evaluated on the same $\mathcal D$.
This yields paired comparisons free of the rollout variation that arises when each drafter is evaluated on independently generated outputs.

\subsubsection{Rounds Committing at Least \texorpdfstring{$k$}{k} Tokens and Acceptance Rate}

A finer-grained statistic is the number of rounds whose \emph{committed length}, i.e., the number of tokens committed in the round including the correction or bonus token, is at least $k$.
Let $\cT^q_{\ge k}$ denote this number in a speculative decoding run.
Since a round commits at most $B+1$ tokens, we consider $1\le k\le B+1$.
A round starting after position $t$ commits at least $k$ tokens if and only if its first $k-1$ drafts are accepted, i.e., it visits state $(t,t+k)$.
This observation generalizes Proposition~\ref{prop:pathwise-rounds} as follows; see Appendix~\ref{sec:proof-pathwise-rounds} for the proof.

\begin{proposition}
\label{prop:pathwise-fine-grained-rounds}
For any target-supported sequence $x=(x_{1:L},\eos)$, any drafter $q$, and any $1\le k\le B+1$,
the expected number of rounds committing at least $k$ tokens to produce $x$ is
\begin{equation}
    \tau_{\ge k}^{q}(x)
    :=
    \mathbb E\big[\cT^q_{\ge k}\,\big|\,X=x\big]
    =
    \sum_{n=0}^{L+1-k}\omega_{n,n+k}^{q}(x_{<n+k}).
    \label{eq:offline-rounds-k}
\end{equation}
\end{proposition}

Taking $k=1$ recovers Proposition~\ref{prop:pathwise-rounds}, so that $\tau_{\mathrm{off}}^{q}=\tau_{\ge 1}^{q}$.
For $1\le k\le B$, we define the acceptance rate at draft position $k$ as
\begin{equation}
    \mathrm{AR}_k(q)
    :=
    \frac{\mathbb E[\cT^q_{\ge k+1}]}{\mathbb E[\cT^q_{\ge k}]},
    \label{eq:acceptance-rate}
\end{equation}
the fraction of rounds reaching draft position $k$ in which the $k$-th draft is also accepted.\footnote{A round whose $k$-th position produces the terminal $\eos$ is counted in $\cT^q_{\ge k}$ but not in $\cT^q_{\ge k+1}$, since generation stops there.}
Using the same dataset $\mathcal D$, we estimate these quantities by
\begin{equation}
    \widehat{\cT}_{\ge k,\mathcal D}(q)
    :=
    \frac1N \sum_{i=1}^{N}\tau_{\ge k}^{q}(x^{(i)})
    \qquad\text{and}\qquad
    \widehat{\mathrm{AR}}_{k,\mathcal D}(q)
    :=
    \frac{\widehat{\cT}_{\ge k+1,\mathcal D}(q)}{\widehat{\cT}_{\ge k,\mathcal D}(q)}.
    \label{eq:offline-acceptance-rate}
\end{equation}
As before, $\widehat{\cT}_{\ge k,\mathcal D}(q)$ is unbiased for $\mathbb E[\cT^q_{\ge k}]$, and $\widehat{\mathrm{AR}}_{k,\mathcal D}(q)$ is consistent for $\mathrm{AR}_k(q)$.

\subsection{Suboptimality of Block-Local Surrogates}
\label{sec:separation}

Finally, we show that maximizing accepted length independently within each block can be strictly suboptimal for the global MAL.
Since maximizing MAL is equivalent to minimizing the expected number of decoding rounds (see Section \ref{sec:notation}), which EDR equals exactly
(Theorem~\ref{thm:exact-edr}), this theoretically justifies optimizing the EDR objective rather than block-local surrogates.

We consider semi-AR drafters that condition on earlier drafts in the block only through the immediately preceding token, i.e.,
$     q_{n,t} ( \cdot \mid x_{1:n};x_{n+1:t-1} ) = q_{n,t} ( \cdot \mid x_{1:n};x_{t-1} ) $,
referred to as \emph{first-order drafters}. This class includes existing semi-AR drafters such as DSpark \citep{dspark} and DFly \citep{angelspec}.
%

The following theorem shows that even when the expected accepted length is optimized in every round, the resulting MAL can be a constant factor below the best achievable within the same drafter class.
The construction and proof are deferred to Appendix~\ref{app:separation}.
\begin{theorem}
\label{thm:separation}
For any block size \(B \geq 3\), there exists a target distribution \(p\) such that, 
for any first-order drafter \(\hat q\) that maximizes the expected accepted length in every round, there exists another first-order drafter \(q^*\) satisfying
\[
    \frac{\MAL(q^*)}{\MAL(\hat q)}
    >
    1+\frac{1}{3B+9}.
\]
\end{theorem}

The gap arises from the cross-round interaction ignored by block-local objectives. 
The draft distribution in the current round determines not only how many tokens are accepted in that round, but also where the next round starts. Consequently, a drafter that is locally optimal for the current round can induce a less favorable distribution of future round starts and result in a lower global MAL.

%% file: related.tex
\section{Related Work}

\subsection{Literature Review}
\paragraph{Drafter architectures for speculative decoding.}
The latency per generated token depends on the accepted length per
round as well as the costs of drafting and verification.
Architectural work therefore seeks to increase accepted length while
keeping drafting inexpensive.
Medusa \citep{cai2024medusa} attaches multiple prediction heads to
target-model features, while Hydra \citep{ankner2024hydra} introduces
sequential dependencies between these heads to improve draft accuracy.
EAGLE-3 \citep{li2026eagle} combines features from multiple target layers. 
To reduce sequential drafting overhead, parallel and diffusion-based
drafters generate multiple proposals in a single forward pass.
DFlash \citep{dflash} uses a lightweight block-diffusion drafter
conditioned on target-model context features.
DiffuSpec \citep{li2026diffuspec} repurposes a pretrained diffusion language model as
a parallel drafter and uses causal-consistency path search to
construct proposals for autoregressive verification.
To improve within-block consistency without returning to fully
autoregressive drafting, DSpark \citep{dspark} and
DFly \citep{angelspec} combine parallel backbones with lightweight
sequential heads that condition on preceding draft tokens.

\paragraph{Training objectives for speculative drafters.}
Drafter training commonly uses cross-entropy or KL-divergence
objectives to align draft predictions with the target distribution
\citep{li2026eagle, dflash}.
However, distributional matching under these losses does not
necessarily maximize token acceptance when drafter capacity is limited.
LK losses \citep{samarin2026lk} address this distinction through
acceptance-oriented objectives. 
For multi-token drafting, the first rejection invalidates all
subsequent proposals in the block, making the contributions of
different positions unequal.
DFlash \citep{dflash} accounts for this asymmetry using fixed
exponentially decaying position weights.
D-PACE \citep{dpace} replaces this fixed schedule with adaptive weights
derived from a differentiable accepted-length surrogate based on
draft confidences.
The end-to-end (E2E) multi-step TV loss \citep{e2e} moves beyond
independent token-level matching by incorporating the multiplicative
structure of acceptance across positions within a block.
Recent work addresses the mismatch between drafter training and
inference by incorporating verification feedback.
Draft-OPD \citep{lei2026draft} replays draft proposals from
inference-time anchors and applies acceptance-dependent
distillation losses.
Verification-Aware Training \citep{gu2026verification} uses
simulated verification outcomes to supervise an auxiliary head
and adapt position weights.
Although these methods incorporate inference-time behavior into
training, their objectives remain surrogates for MAL.


\paragraph{Diffusion language models.}
Diffusion language models (DLMs) \citep{nie2025large,labs2025mercury,ye2025dream} offer an alternative to AR generation by iteratively refining sequences with bidirectional context, enabling parallel token generation.
Continuous DLMs \citep{li2022diffusion,chen2026langflow,li2026convergeflow} extend continuous diffusion models \citep{ho2020denoising,song2020score,song2020denoising} and flow matching \citep{lipman2022flow} to token embedding spaces. Starting from Gaussian noise, they generate continuous token representations through learned denoising or flow dynamics, then map these representations to discrete tokens. This framework facilitates adapting established techniques for controlling and accelerating diffusion sampling, including classifier-free guidance \citep{ho2022classifier}, self-conditioning \citep{chen2022analog}, few-step solvers \citep{lu2022dpm,li2024provable,jiao2026first}, and distillation \citep{song2024improved,yin2024one}.
Discrete DLMs \citep{lou2023discrete,sahoo2024simple,shi2024simplified} operate directly in token space using categorical corruption. 
Masked DLMs belong to this category and corrupt sequences by replacing tokens with a special mask token. Then it generate text by progressively unmasking multiple positions in parallel. This formulation underlies many parallel and semi-AR drafters.
However, conditionally independent sampling within an update can overlook dependencies among simultaneously generated tokens, creating a trade-off between generation quality and sampling efficiency \citep{li2025breaking,chen2025optimal,dmitriev2026efficient,zhao2026adaptation,cai2026confidence}.

%% file: loss.tex
\subsection{Comparison of Training Objectives}
\label{app:loss-formulations}

In this section, we present a unified summary of existing drafter training objectives in the literature, together with our proposed EDR objective.

Let $X=(X_1,\dots,X_L,\eos)$ be a rollout sequence sampled from the target distribution, and let $B$ be the number of proposed draft tokens in a block.
The block is drafted from the committed prefix $X_{1:n}$; $q_{n,t}$ and $p_t$ denote the draft and target distributions at position $t$, respectively.
We suppress their dependence on the sampled sequence for brevity.

Up to additive constants and positive scaling factors, the objectives considered here can be written as
\begin{equation}
    \mathcal L(q)
    =
    \E_{X\sim p}\Bigg[\sum_{n=0}^{L} a_n \sum_{t=n+1}^{\min\{n+B+1, L\}} b_{n,t}\,\ell_{n,t}\Bigg].
    \label{eq:loss-template}
\end{equation}
This template separates each objective into three components:
\begin{itemize}
    \item \emph{Per-position loss} $\ell_{n,t}$ specifies the local training signal at position $t$ for a block drafted from anchor $n$.
    For token-level alignment surrogates, $\ell_{n,t}$ measures the discrepancy between the draft distribution $q_{n,t}$ and the target $p_t$, whereas for acceptance-aware objectives, $\ell_{n,t}$ is the surrogate for the rejection probability at position $t$.
    All objectives except EDR set $\ell_{n,n+B+1}=0$, since no draft token is proposed at the bonus position.
    EDR keeps this term because a round that reaches the bonus position always ends there ($A^q_{n,n+B+1}=0$).
    \item \emph{Block position weight} $b_{n,t}$ sets how much position $t$ counts relative to the other positions in the same block.
    Later draft tokens are verified only if all earlier draft tokens in the block are accepted, so their contributions depend on acceptance at preceding positions.
    Non-uniform position weights $b_{n,t}$ reflect this asymmetry between positions.
    \item \emph{Anchor weight} $a_n$ sets how much the block drafted from anchor $n$ counts relative to the blocks drafted from other anchors.
    All objectives except EDR weight anchors equally, i.e., $a_n=1$.
\end{itemize}
Table~\ref{tab:loss-summary} summarizes the three components of each objective. Here, $\sg(\cdot)$ denotes the stop-gradient operator: $\sg(z)$ has the same value as $z$, but it is treated as a constant when taking gradients with respect to the drafter parameters.

We discuss these three groups below.


\begin{table}[t]
\centering
\small
\setcellgapes{4pt}
\makegapedcells
\renewcommand{\cellalign}{c}
\caption{Drafter training objectives as instances of the template in~\eqref{eq:loss-template}.}
\label{tab:loss-summary}
\begin{tabular}{cccc}
\toprule
Objective & Per-position loss $\ell_{n,t}$ & Block position weight $b_{n,t}$ & Anchor weight $a_n$ \\
\midrule
\multicolumn{4}{l}{\emph{Token-level surrogates}} \\
CE & $-\log q_{n,t}(x_t)$ & $1$ & $1$ \\
KL & $\KL(p_t\,\|\,q_{n,t})$ & $1$ & $1$ \\
TV & $\TV(p_t,q_{n,t})$ & $1$ & $1$ \\
\makecell{LK \\ \citep{samarin2026lk}}
    & \makecell{$\big(\lambda\,\KL+(1-\lambda)\,\TV\big)$, \\
               $\lambda=\exp\bigl(-\kappa\sg(1-\TV)\bigr)$}
    & $1$ & $1$ \\
\midrule
\multicolumn{4}{l}{\emph{Block-level surrogates}} \\
\makecell{Exponential decay \\ \citep{dflash}}
    & \makecell{CE / KL / TV / LK \\ or a weighted combination}
    & $\exp\bigl(-(t-n-1)/\eta\bigr)$
    & $1$ \\
\makecell{D-PACE \\ \citep{dpace}}
    & \makecell{CE / KL / TV / LK \\ or a weighted combination}
    & \makecell{$\displaystyle \sg\Big(\sum_{m=t-n}^{B}\,\prod_{s=n+1}^{n+m} \widetilde{q}_{n, s}(x_s) \Big)$, \\ $ \widetilde{q}_{n, s}  = \big(1 - \alpha \big) q_{n,s}+ \alpha $}
    & $1$ \\
\makecell{E2E-TV \\ \citep{e2e}}
    & $(\TV(p_t,q_{n,t})-1)$
    & $\displaystyle \prod_{s=n+1}^{t-1}\big(1-\TV(p_s,q_{n,s})\big)$
    & $1$ \\
\midrule
\multicolumn{4}{l}{\emph{Exact objective}} \\
\makecell{\textbf{EDR} \\ (ours)}
    & $c^q_{n,t}$
    & $\displaystyle \prod_{s=n+1}^{t-1}A^q_{n,s}$
    & $\omega^q_{n,n+1}$ \\
\makecell{\textbf{EDR (TD form)} \\ (ours)}
    & $c^q_{n,t}+A^q_{n,t}\,\sg\big(V^q_{n,t+1}-V^q_{t,t+1}\big)$
    & $\displaystyle \sg\Big(\prod_{s=n+1}^{t-1}A^q_{n,s}\Big)$
    & $\sg(\omega^q_{n,n+1})$ \\
\bottomrule
\end{tabular}
\end{table}

\paragraph{Token-level surrogates.}
Token-level surrogates set $b_{n,t}=a_n=1$ and align the draft distribution $q_{n,t}$ with the target $p_t$ at each position $t$ independently.
CE and KL both minimize a KL divergence between a reference distribution and $q_{n,t}$:
the reference is the one-hot distribution at the sampled target token $x_t$ for CE and the full target distribution $p_t$ for KL.
TV uses the TV distance, which  equals the rejection probability at position $t$.
LK interpolates between KL and TV using an acceptance-dependent coefficient $\lambda$ that balances acceptance awareness against ease of optimization.
As the acceptance probability $1-\TV(p_t,q_{n,t})$ increases, $\lambda$ decreases, so LK relies mainly on the KL term that is easier to optimize when acceptance is low and moves weight to the TV term as acceptance improves.

\paragraph{Block-level surrogates.}
Block-level surrogates extend token-level surrogates with nonuniform block position weights $b_{n,t}$, which reflect the asymmetry between positions within a block.
Later positions receive smaller weights because they are verified less often---a position is verified only if all earlier draft tokens in its block are accepted.

Exponential decay and D-PACE multiply a token-level loss by a weight under stop-gradient. The weight is a fixed schedule in exponential decay and a confidence-based surrogate of the accepted length in D-PACE.
E2E uses $\ell_{n,t}=\TV(p_t,q_{n,t})-1$ at draft positions and $b_{n,t}=\prod_{s=n+1}^{t-1}(1-\TV(p_s,q_{n,s}))$ without stop-gradient.
Therefore, the inner sum in~\eqref{eq:loss-template} is given by
\begin{equation*}
    -\sum_{k=1}^{\min\{B,L+1-n\}}\prod_{s=n+1}^{n+k}\big(1-\TV(p_s,q_{n,s})\big),
\end{equation*}
the negative of a surrogate for the expected accepted length of the block drafted from anchor $n$.
In contrast, the block position weight of EDR is the conditional survival probability $b_{n,t}=\prod_{s=n+1}^{t-1}A^q_{n,s}$ along the target rollout. It is exact in expectation, whereas the product of $1-\TV$ terms in E2E is not.

Like token-level surrogates, block-level surrogates ignore cross-round coupling, treating each block separately and assigning the same weight $a_n=1$ to all anchors.

\paragraph{EDR.}
For EDR, the three components are the EOS-aware local rejection cost $\ell_{n,t}=c^q_{n,t}$, the within-round survival probability $b_{n,t}=\prod_{s=n+1}^{t-1}A^q_{n,s}$, and the anchor weight $a_n=\omega^q_{n,n+1}$.
By the recursion in \eqref{eq:occupancy-continue}, we have
\begin{equation*}
    a_n\,b_{n,t}
    =
    \omega^q_{n,n+1}\prod_{s=n+1}^{t-1}A^q_{n,s}
    =
    \omega^q_{n,t},
\end{equation*}
which is the probability, conditioned on the target rollout,  that decoding visits state $(n,t)$.
Thus, with these choices, the weighted sum inside the expectation in~\eqref{eq:loss-template}, plus the constant $1$, equals the per-sequence EDR loss~\eqref{eq:edr-training-estimator}.
Taking the expectation over target rollouts gives the expected number of decoding rounds.

Among the objectives in Table~\ref{tab:loss-summary}, only EDR has drafter-dependent anchor weights.
These weights capture how acceptance decisions change where later rounds start, that is, the cross-round coupling that the surrogates above ignore.

\paragraph{EDR in TD form.}
The TD form of EDR is a different scalar loss, but it has the same gradient with respect to the drafter parameters as the EDR loss above (see Theorem~\ref{prop:edr-gradient}).
It applies stop-gradient to both the block position weights and the anchor weights, and it adds the term $A^q_{n,t}\,\sg\big(V^q_{n,t+1}-V^q_{t,t+1}\big)$ to the per-position loss.
Gradients therefore flow only through $c^q_{n,t}$ and $A^q_{n,t}$, not through $\omega^q$ or $V^q$, so training with the TD form is more efficient.
We use this form in training (see Algorithm~\ref{alg:edr}).



%% file: experiments.tex
\section{Experiments}
\label{sec:experiments}

We evaluate whether directly optimizing EDR improves self-speculative decoding over existing block-local training objectives.  
We consider two SOTA self-speculative drafters,
DSpark~\citep{dspark} and DFly~\citep{angelspec}, paired with Qwen3 \citep{qwen3} target models of different scales.

\subsection{Experimental Setup}
\label{sec:experimental-setup}

\paragraph{Models and sampling.}
We evaluate two target--drafter pairs: Qwen3-4B with DSpark and Qwen3-8B
with DFly. We set block size $ B=7 $ for both drafters. For Qwen3-4B, target trajectories are sampled with
temperature $0.7$, top-$p=0.8$, and top-$k=20$, while the DSpark proposal
distribution uses temperature $0.7$ without top-$p$ or top-$k$ truncation.
For Qwen3-8B, both the target and DFly drafter use temperature $1.0$
without top-$p$ or top-$k$ truncation. Thinking mode is disable in all settings.

\paragraph{Training objectives.}
Starting from the corresponding publicly released draft-model checkpoint, we finetune
only the drafter parameters.  We compare our exact EDR objective against the
end-to-end TV (E2E) block-local objective~\citep{e2e}, using the same model
architecture, training data, optimizer configuration, and number of update
steps for both objectives. For all methods, we use at most 512 training anchors per trajectory.  
For EDR, we use importance sampling to select training anchors. For details, see Appendix \ref{app:importance-sampling}. 

\paragraph{Training data.}
We train on Open-PerfectBlend \citep{OPB}.  Rather than using reference answers, we
generate a fresh trajectory for each prompt from the target model using the same target sampling configuration as evaluation.  
Each drafter is finetuned for one epoch with batch size $96$.

\paragraph{Evaluation.}
We use the same evaluation prompts as DSpark~\citep{dspark} on nine
benchmarks spanning math \citep{GSM8K, math500, aime25}, code \citep{mbpp, humaneval, LCB}, and chat \citep{MT-Bench, alpaca, arena-hard}.  
We generate at least $1{,}500$
target trajectories per benchmark (multiple trajectories for each prompt) under the same sampling configurations as in training.
We compute MAL using the offline estimator introduced in
Section~\ref{sec:offline-evaluation}. 
All methods for the same target model and sampling configurations are evaluated on exactly the same target trajectories.
This provides paired comparisons across training objectives and removes additional
variance from independently sampled evaluation outputs.

All training and evaluation are performed on a single NVIDIA H200 GPU or H100 GPU.  
For more experimental details, see Appendix \ref{app:exp-details}.

\subsection{Experimental Results}

\begin{table}[t]
    \centering
    \caption{
        \textbf{One-epoch finetuning results.} 
        We report mean accepted length (MAL; higher is better)
        after one epoch of draft-model finetuning.
        Target and draft sampling configurations in evaluation are the same as in training.
        Block size $B=7$ for all drafters.
        \textbf{Bold} marks the best finetuned result.
    }
    \label{tab:main-mal}
    \resizebox{\textwidth}{!}{
        \setlength{\tabcolsep}{2pt}
        \renewcommand{\arraystretch}{1.08}
        \begin{tabular}{@{}llcccccccccc@{}}
            \toprule
            \multirow{2}{*}{Target}
            & \multirow{2}{*}{ \raisebox{-3.5ex}{\shortstack[c]{
            Drafter\\
            {($B=7$)}
            }}}
            & \multicolumn{3}{c}{Math}
            & \multicolumn{3}{c}{Code}
            & \multicolumn{3}{c}{Chat} \\
            \cmidrule(lr){3-5}
            \cmidrule(lr){6-8}
            \cmidrule(lr){9-11}
            &
            & GSM8K & MATH & AIME25
            & MBPP & HumanEval & LCB
            & MT-Bench & Alpaca & Arena-Hard
            & \\
            \midrule

            \multirow{3}{*}{
        \shortstack[l]{
            Qwen3-4B\\
            {\scriptsize ($T=0.7$}\\
            {\scriptsize top-$p=0.8$}\\
            {\scriptsize top-$k=20$)}
        }
    }
            & DSpark
            & 6.23
            & 6.12
            & 5.46
            & 5.28
            & 5.52
            & 5.46
            & 3.70
            & 3.62
            & 3.59
            \\

            & E2E
            & \textbf{6.32}
            & \textbf{6.20}
            & 5.57
            & \textbf{5.34}   
            & 5.58
            & 5.54
            & 3.75
            & 3.66
            & 3.67
            \\

            & \textbf{EDR (Ours)}
            & \textbf{6.32}
            & \textbf{6.20}
            & \textbf{5.59}
            & \textbf{5.34}
            & \textbf{5.59}
            & \textbf{5.56}
            & \textbf{3.78}
            & \textbf{3.70}
            & \textbf{3.70}
            \\

            \midrule

            \multirow{3}{*}{
                \shortstack[l]{
                    Qwen3-8B\\
                    {\scriptsize ($T=1.0$}\\
                    {\scriptsize no top-$p$/top-$k$)}
                }
            }
            & DFly
            & 6.30
            & 5.94
            & 5.15
            & 5.26
            & 5.58
            & 5.23
            & 3.57
            & 3.45
            & 3.18
            \\

            & E2E
            & 6.34
            & 5.98
            & 5.18
            & \textbf{5.32}
            & 5.62
            & 5.26
            & 3.61
            & 3.49
            & 3.21
            \\

            & \textbf{EDR (Ours)}
            & \textbf{6.35}
            & \textbf{6.00}
            & \textbf{5.22}
            & \textbf{5.32}
            & \textbf{5.63}
            & \textbf{5.29}
            & \textbf{3.65}
            & \textbf{3.52}
            & \textbf{3.25}
            \\
            \bottomrule
        \end{tabular}
    }
\end{table}

Table~\ref{tab:main-mal} shows that EDR consistently improves speculative decoding across both DSpark and DFly without modifying the drafter architecture. 
Compared with E2E finetuning, EDR achieves higher or equal MAL across all benchmarks, with larger gains on AIME25, LiveCodeBench, and the chat benchmarks.
These results suggest that directly optimizing the global expected number of decoding rounds provides additional gains beyond block-local end-to-end surrogates, and that the improvement transfers across different semi-AR drafter architectures and target-model scales.

%% file: conclusion.tex
\section{Conclusion}
\label{sec:conclusion}
We introduced Expected Decoding Rounds (EDR), an exact global training objective for parallel and semi-AR speculative drafters. 
By formulating speculative decoding as a Markov reward process conditioned on target-model outputs, we derived a tractable gradient for directly minimizing the expected number of verification rounds, together with an exact offline evaluator for paired draft-model comparison. 
Empirically, EDR consistently improves two strong drafters, DSpark and DFly, across nine benchmarks without modifying their architectures or inference procedures, and outperforms the E2E training objective in overall mean accepted length. 
These results suggest that explicitly accounting for cross-round dependencies provides a principled and practical direction for training more efficient parallel speculative draft models.

We conclude with several directions for future work.
First, our current framework focuses on a fixed draft block size. 
However, under high-concurrency serving, verifying long draft blocks can become inefficient, especially when many later proposals are likely to be rejected. 
Adaptive block sizing can reduce unnecessary verification by shortening the draft block when confidence is low, thereby further reducing inference latency~\citep{hu2026echo, huang2024specdec++}. 
An important direction is to extend our framework to adaptive block sizes and to study whether the optimal block size can be selected dynamically for each decoding round.
%
Second, EDR requires computing occupancy weights $\omega$ and value functions $V$ for multiple anchor positions. 
This is more computationally expensive than block-local objectives. 
It would be interesting to explore more efficient ways to utilize the occupancy and value computations. 
%
Finally, as shown in Table~\ref{tab:main-mal}, MAL varies substantially across datasets, suggesting that some target distributions are intrinsically more difficult to draft than others. 
Our current work focuses on optimizing the drafter given a target distribution, but does not provide a complete theoretical characterization of this difficulty. 
A promising direction is to identify properties of the target distribution that determine the best achievable speculative decoding efficiency.

%% file: algorithm.tex
\section{Speculative Decoding Algorithm}
\label{sec:algorithm}
Algorithm~\ref{alg:self-sd} summarizes the speculative decoding procedure we consider.

\begin{algorithm}[h]
\small
\caption{Speculative decoding with a semi-autoregressive drafter}
\label{alg:self-sd}
\begin{algorithmic}[1]
\Require Block size $B$, target distributions $p_t$, draft distributions $q_{n,t}$
\State $n\gets0$, $\cT\gets0$
\Comment{$n$: number of committed ordinary tokens}

\While{\textbf{true}}
    \State Draw a draft block
    $\widetilde x_{n+1:n+B}\sim q_{n, n+1}(\cdot\mid x_{1:n}) \cdots q_{n, n+B}(\cdot \mid x_{1:n}; \widetilde x_{n+1:n+B-1} )$
    \State Evaluate
    $p_t(\cdot\mid x_{1:n},\widetilde x_{n+1:t-1})$
    for $t=n+1,\ldots,n+B+1$ in one target-model pass
    \State $\cT\gets\cT+1$
    \State $\mathrm{rejected}\gets\mathrm{false}$

    \For{$t=n+1,\ldots,n+B$}
        \State
        $p\gets p_t(\cdot\mid x_{1:n},\widetilde x_{n+1:t-1})$
        \State
        $q\gets q_{n,t}(\cdot\mid x_{1:n};\widetilde x_{n+1:t-1})$
        \State Draw $U_t\sim\mathrm{Unif}[0,1]$

        \If{$U_t\le
            \min\{1,p(\widetilde x_t)/q(\widetilde x_t)\}$}
            \State $x_t\gets\widetilde x_t$
        \Else
            \State Draw $ x_t\sim[p-q]_+ / \TV(p, q)$
            \State $\mathrm{rejected}\gets\mathrm{true}$
        \EndIf

        \If{$x_t=\eos$}
            \State \Return $(x_{1:t-1}, \eos)$ and $\cT$
        \EndIf

        \If{$\mathrm{rejected}$}
            \State $n\gets t$
            \State \textbf{break}
        \EndIf
    \EndFor

    \If{\textbf{not} $\mathrm{rejected}$}
        \State Draw
        $\widetilde x_{n+B+1}\sim p_{n+B+1}(\cdot\mid x_{1:n+B})$
        \Comment{bonus token}

        \If{$\widetilde x_{n+B+1}=\eos$}
            \State \Return $(x_{1:n+B}, \eos)$ and $\cT$
        \Else
            \State $x_{n+B+1}\gets \widetilde x_{n+B+1}$
            \State $n\gets n+B+1$
        \EndIf
    \EndIf
\EndWhile
\end{algorithmic}
\end{algorithm}

%% file: proof.tex
\section{Proofs}
\subsection{Markov Process Formulation in Section \ref{sec:pathwise-transition}}
\label{sec:markov-chain}
In this section, we give a detailed derivation of the rejection-sampling
transition dynamics conditioned on a fixed target-supported output
sequence $x=(x_{1:L},\texttt{<EOS>})$.
Specifically, we define the Markov chain states, derive the conditional
transition probabilities, and prove that the occupancy
$\omega_{n,t}^q(x)$ depends on $x$ only through $x_{<t}$.
Throughout, all probabilities refer to speculative decoding with a
fixed drafter $q$.

\paragraph{States and decoding histories.}
For each position $t$ reached by decoding, let $N_t$ denote the
round-start index of the round that produces $X_t$.
Thus, immediately before processing position $t$, the state is
\[
    S_t=(N_t,t),
    \qquad S_1=(0,1).
\]
If $N_t=n$, then the current round started from the committed prefix
$X_{1:n}$, and all preceding drafts in this round have been accepted.
Consequently, their values equal $X_{n+1:t-1}$.
For $t\le n+B$, position $t$ is processed by draft verification;
for $t=n+B+1$, it is processed by drawing a target bonus token.
States with $t>n+B+1$ are unreachable.

Let
\[
    \mathcal H_t
    :=\sigma(X_{<t},N_1,\ldots,N_t)
\]
denote the history of committed tokens and round-start indices before
processing position $t$.
We first analyze the next output token conditional on this history,
without conditioning on the complete output sequence.

\paragraph{The local rejection-sampling law.}
Consider a history with $X_{<t}=x_{<t}$ and $N_t=n$ at a draft
position $t\le n+B$.
For brevity, write
\[
    p_t(v):=p_t(v\mid x_{<t}),
    \qquad
    q_t(v):=q_{n,t}(v\mid x_{1:n};x_{n+1:t-1}),
    \qquad v\in\overline{\mathcal V}.
\]
Let $\mathrm{acc}_t$ and $\mathrm{rej}_t$ denote the acceptance
and rejection events at this position.
The verification rule gives
\begin{align*}
    \Pr(X_t=v,\mathrm{acc}_t\mid\mathcal H_t)
    &=\min\{p_t(v),q_t(v)\},\\
    \Pr(X_t=v,\mathrm{rej}_t\mid\mathcal H_t)
    &=[p_t(v)-q_t(v)]_+.
\end{align*}
The first identity follows by multiplying the proposal probability by
the verifier's acceptance probability.
For the second identity, rejection has total probability
$\operatorname{TV}(p_t,q_t)$, and the correction token is drawn from
the residual distribution
$[p_t-q_t]_+/\operatorname{TV}(p_t,q_t)$.
When $\operatorname{TV}(p_t,q_t)=0$, rejection has probability zero
and no residual draw is needed.

Adding the two identities yields
\begin{equation}
\label{eq:markov-next-token-law}
    \Pr(X_t=v\mid\mathcal H_t)
    =
    \min\{p_t(v),q_t(v)\}+[p_t(v)-q_t(v)]_+
    =
    p_t(v\mid X_{<t}).
\end{equation}
At a bonus-token position, the same identity holds because the token
is drawn directly from the target distribution.
Therefore, conditional on any decoding history, the next committed
token has the target conditional distribution, regardless of the
current round-start index.

\paragraph{Transition kernels conditioning on the complete output.}
Equation~\eqref{eq:markov-next-token-law} implies that, on any history
with $X_{<t}=x_{<t}$,
\begin{equation}
\label{eq:markov-continuation-law}
    \Pr(X=x\mid\mathcal H_t)
    =
    p(x_{t:L+1}\mid x_{<t})
    :=
    \prod_{s=t}^{L+1}p_s(x_s\mid x_{<s}).
\end{equation}
To see this rigorously, use backward induction along the fixed
sequence $x$.
At $t=L+1$, the required remaining output is just
$\texttt{<EOS>}$, so the identity follows directly from
\eqref{eq:markov-next-token-law}.
Suppose it holds at $t+1$ for every possible decoding history.
After committing $X_t=x_t$, every possible resulting history has
the same remaining-output probability
$p(x_{t+1:L+1}\mid x_{\le t})$.
Multiplying this probability by
$\Pr(X_t=x_t\mid\mathcal H_t)=p_t(x_t\mid x_{<t})$
proves the identity at $t$.

In particular, after committing the same non-EOS token $x_t$,
the future suffix has the same probability whether the token was
accepted or produced by residual sampling.
Thus, for $t\le L$ at a draft position, Bayes' rule gives
\begin{align*}
    \Pr(\mathrm{acc}_t\mid\mathcal H_t,X=x)
    &=
    \frac{
        \min\{p_t(x_t),q_t(x_t)\}\,
        p(x_{t+1:L+1}\mid x_{\le t})
    }{
        p_t(x_t)\,
        p(x_{t+1:L+1}\mid x_{\le t})
    }\\
    &=
    \frac{\min\{p_t(x_t),q_t(x_t)\}}{p_t(x_t)}\\
    &=
    \min\left\{
        1,
        \frac{
            q_{n,t}(x_t\mid x_{1:n};x_{n+1:t-1})
        }{
            p_t(x_t\mid x_{<t})
        }
    \right\}\\
    &=A_{n,t}^q(x_{\le t}).
\end{align*}
All target-probability factors being cancelled are positive because
$x$ is target-supported.
Similarly,
\[
    \Pr(\mathrm{rej}_t\mid\mathcal H_t,X=x)
    =
    1-A_{n,t}^q(x_{\le t})
    =
    R_{n,t}^q(x_{\le t}).
\]
Notice that the ratio here is $q_t(x_t)/p_t(x_t)$, rather than
the verifier's ratio $p_t(\widetilde X_t)/q_t(\widetilde X_t)$:
we are conditioning on the committed output token, not on the
proposed token.

If the token is accepted, the current round continues.
If it is rejected, the next round starts after position $t$.
At a bonus-token position $t=n+B+1$, the next round also starts
after position $t$, provided that the bonus token is not EOS.
Accordingly, setting
\[
    A_{n,t}^q=0,
    \qquad
    R_{n,t}^q=1,
    \qquad t>n+B,
\]
incorporates the bonus-token case into the same transition rule.
At such a position, $R_{n,t}^q=1$ represents a forced round transition,
not an actual draft rejection.

It follows that, under $\Pr(\,\cdot\mid X=x)$, the transitions
for $t\le L$ are
\[
    (n,t)\longrightarrow
    \begin{cases}
        (n,t+1),
        &\text{with probability }A_{n,t}^q(x_{\le t}),\\
        (t,t+1),
        &\text{with probability }R_{n,t}^q(x_{\le t}).
    \end{cases}
\]
With $x$ fixed, these probabilities depend on the past history only
through the current state $(n,t)$.
This establishes the Markov property.

The conditional chain has states
$0\le n<t\le L+1$ and is stopped at position $L+1$.
All states $(n,L+1)$ are terminal for the round-count process:
the final EOS is produced in the current round and does not start
another round, regardless of whether it is accepted, sampled as a
correction, or sampled as a bonus token.

\paragraph{Occupancy depends only on the committed prefix.}
Since the position index increases by one at every transition,
state $(n,t)$ is visited exactly when $N_t=n$.
Hence,
\[
    \omega_{n,t}^q(x)
    =
    \Pr(N_t=n\mid X=x).
\]
By \eqref{eq:markov-continuation-law}, conditional on the committed
prefix $x_{<t}$, the probability of the remaining output
$x_{t:L+1}$ is the same for every possible round-start history.
Therefore,
\begin{align*}
    \omega_{n,t}^q(x)
    &=
    \frac{
        \Pr(N_t=n,X_{<t}=x_{<t})\,
        p(x_{t:L+1}\mid x_{<t})
    }{
        \Pr(X_{<t}=x_{<t})\,
        p(x_{t:L+1}\mid x_{<t})
    }\\
    &=
    \Pr(N_t=n\mid X_{<t}=x_{<t})\\
    &=:\omega_{n,t}^q(x_{<t}).
\end{align*}
This proves that the occupancy depends only on $x_{<t}$,
even though it was originally defined by conditioning on the
complete output sequence.

Starting from $\omega_{0,1}^q=1$, the conditional Markov transitions
give the forward recursions
\begin{align*}
    \omega_{n,t+1}^q(x_{\le t})
    &=
    \omega_{n,t}^q(x_{<t})A_{n,t}^q(x_{\le t}),
    &&0\le n<t\le L,\\
    \omega_{t,t+1}^q(x_{\le t})
    &=
    \sum_{n=0}^{t-1}
    \omega_{n,t}^q(x_{<t})R_{n,t}^q(x_{\le t}),
    &&1\le t\le L.
\end{align*}
The first recursion accounts for continuing the current round.
The second sums over all states from which a new round can start
after position $t$.
These are precisely the occupancy recursions used in
Section~\ref{sec:pathwise-transition}.

\subsection{Proofs of Proposition \ref{prop:pathwise-rounds} and Proposition \ref{prop:pathwise-fine-grained-rounds}}
\label{sec:proof-pathwise-rounds}
We first prove Proposition~\ref{prop:pathwise-fine-grained-rounds}, and then obtain Proposition~\ref{prop:pathwise-rounds} as its special case $k=1$.

\paragraph{Proof of Proposition~\ref{prop:pathwise-fine-grained-rounds}.}
Fix $1\le k\le B+1$.
The random variable \(\cT^q_{\ge k}\) counts the decoding rounds that commit at least $k$ tokens.
Conditioned on a fixed output sequence \(x\), consider the round that starts after position \(n\), i.e., from the committed prefix $x_{1:n}$.
By the transition rule~\eqref{eq:mrp-transition}, a rejection transition always leads to a diagonal state, so a non-diagonal state $(n,t)$ with $t>n+1$ can only be reached from $(n,t-1)$ by an acceptance.
Hence state \((n,n+k)\) is visited exactly when this round has started and its drafts at positions $n+1,\ldots,n+k-1$ have all been accepted.
In that case, position $n+k$ is processed in the same round and commits one more token, which is an accepted draft, a correction token, a bonus token, or the terminal \(\eos\).
Conversely, if the round commits at least $k$ tokens, its first $k-1$ committed tokens must be accepted drafts: a correction token or \(\eos\) ends the round, and a bonus token can only be the $(B+1)$-th committed token.
Therefore, the round starting after position $n$ commits at least $k$ tokens exactly when state $(n,n+k)$ is visited.
It follows that
\begin{align*}
    \cT^q_{\ge k} = \sum_{n=0}^{L+1-k} \mathbf 1\{\text{state } (n,n+k) \text{ is visited}\}.
\end{align*}
Conditioned on $X=x$, the event that state $(n,n+k)$ is visited has probability \(\omega_{n,n+k}^q(x_{<n+k})\). 
Therefore, by linearity of expectation,
\begin{align*}
    \tau_{\ge k}^q(x) = \E[\cT^q_{\ge k} \mid X = x] = \sum_{n=0}^{L+1-k} \omega_{n, n+k}^q(x_{<n+k}).
\end{align*}

\paragraph{Proof of Proposition~\ref{prop:pathwise-rounds}.}
Every round commits at least one token, so $\cT^q=\cT^q_{\ge 1}$.
Equivalently, a round starts after position \(t\) exactly when state \((t,t+1)\) is visited.
Taking $k=1$ in Proposition~\ref{prop:pathwise-fine-grained-rounds} gives
\begin{align*}
    \tau^q(x) = \E[\cT^q \mid X = x] = \sum_{t=0}^{L} \omega_{t, t+1}^q(x_{\leq t}).
\end{align*}
Separating the term $t=0$, using $\omega_{0,1}^q=1$, and substituting the occupancy recursion~\eqref{eq:occupancy-restart} for $1\le t\le L$ gives
\begin{align*}
    \tau^q(x)
    = \omega_{0,1}^q + \sum_{t=1}^{L} \omega_{t, t+1}^q(x_{\leq t})
    = 1+
    \sum_{t=1}^{L}\sum_{n=0}^{t-1}
    \omega_{n,t}^{q}(x_{<t})R_{n,t}^{q}(x_{\leq t}).
\end{align*}

\subsection{Proof of Theorem \ref{thm:exact-edr}}
\label{sec:proof-exact-edr}
To handle the randomness of the sequence length $ L $, 
we rewrite Proposition~\ref{prop:pathwise-rounds} and take the expectation over $ X \sim p_{\target}  $,
\begin{equation}
    \E[\tau^{q}(X)]
    =
    1+
    \sum_{t\ge1}\sum_{n=0}^{t-1}
    \E\!\left[
        \mathbf 1_{\{t\le L\}}
        \omega_{n,t}^{q}(X_{<t})R_{n,t}^{q}(X_{\leq t} )
    \right].
    \label{eq:raw-edr}
\end{equation}
Before $X_t$ is revealed, $\omega_{n,t}^{q}(X_{<t} )$ is determined by $X_{<t}$,
and $t\le L$ is exactly the event $X_t\neq\eos$.  Hence, conditioning on
$X_{<t}$ and using the definition of $c_{n,t}^{q}$,
\begin{align}
    &\E\!\left[
        \mathbf 1_{\{t\le L\}}
        \omega_{n,t}^{q}(X_{<t} )R_{n,t}^{q}(X_{\leq t} )
        \mid X_{<t}
    \right]
    \notag\\
    &\qquad=
    \omega_{n,t}^{q}(X_{<t} )
    \Pr(X_t\neq\eos\mid X_{<t})
    c_{n,t}^{q}(X_{<t})
    \notag\\
    &\qquad=
    \E\!\left[
        \mathbf 1_{\{t\le L\}}
        \omega_{n,t}^{q}(X_{<t} )c_{n,t}^{q}(X_{<t})
        \mid X_{<t}
    \right].
    \label{eq:weighted-cost-rb}
\end{align}
Taking expectations and summing over $n$ and $t$ proves
\eqref{eq:exact-decoding-rounds}.

\subsection{Proof of Theorem \ref{prop:edr-gradient}}
\label{sec:proof-edr-gradient}
    We first prove \eqref{eq:weighted-cost-from-V}, which relates the initial-state value to the per-sequence EDR loss.
\begin{align*}
    V_{0,1}^{q}(x)
    &=
    \sum_{t=1}^{L}
    \left[
        \sum_{n=0}^{t-1}
        \omega_{n,t}^{q}(x_{<t})V_{n,t}^{q}(x)
        -
        \sum_{n=0}^{t}
        \omega_{n,t+1}^{q}(x_{\leq t})V_{n,t+1}^{q}(x)
    \right]
    \\
    &=
    \sum_{t=1}^{L}\sum_{n=0}^{t-1}
    \omega_{n,t}^{q}(x_{<t})
    \left[
        V_{n,t}^{q}(x)
        -
        A_{n,t}^{q}(x_{\leq t})V_{n,t+1}^{q}(x)
        -
        R_{n,t}^{q}(x_{\leq t})V_{t,t+1}^{q}(x)
    \right]
    \\
    &=
    \sum_{t=1}^{L}\sum_{n=0}^{t-1}
    \omega_{n,t}^{q}(x_{<t})c_{n,t}^{q}(x_{<t})
    \\
    &=
    \widehat{\mathcal L}_{\mathrm{EDR}}(q_\theta;x)-1.
\end{align*}

    Now we prove Theorem \ref{prop:edr-gradient}.  Differentiating the Bellman equation \eqref{eq:V-bellman} at one state gives
\begin{align}
    \nabla_\theta V_{n,t}^{q}
    &=
    \underbrace{
    \nabla_\theta c_{n,t}^{q}
    +\nabla_\theta A_{n,t}^{q}
        \left(V_{n,t+1}^{q}-V_{t,t+1}^{q}\right)
    }_{\text{local TD gradient}}
    \notag\\
    &\qquad
    +A_{n,t}^{q}\nabla_\theta V_{n,t+1}^{q}
    +R_{n,t}^{q}\nabla_\theta V_{t,t+1}^{q},
    \label{eq:V-gradient-recursion}
\end{align}
where we used $R_{n,t}^{q}=1-A_{n,t}^{q}$.  Unrolling the last two terms from
the initial state weights each local TD gradient by the probability of
visiting that state, namely $\omega_{n,t}^{q}$.  
Apply \eqref{eq:V-gradient-recursion} recursively from $(0,1)$.  Every time a
state $(n,t)$ is reached, its local term is accumulated once, while the two
recursive terms propagate with transition probabilities $A_{n,t}^{q}$ and
$R_{n,t}^{q}$.  The total coefficient on the local term at $(n,t)$ is
therefore exactly its state occupancy $\omega_{n,t}^{q}$.  Using
\eqref{eq:weighted-cost-from-V} yields \eqref{eq:edr-gradient}.
Rigorously, 
\begin{align*}
    \nabla_\theta\widehat{\mathcal L}_{\mathrm{EDR}}(q_\theta;x)
    &=
    \nabla_\theta V_{0,1}^{q}(x)
    \\
    &=
    \sum_{t=1}^{L}
    \left[
        \sum_{n=0}^{t-1}
        \omega_{n,t}^{q}(x_{<t})\nabla_\theta V_{n,t}^{q}(x)
        -
        \sum_{n=0}^{t}
        \omega_{n,t+1}^{q}(x_{\leq t})\nabla_\theta V_{n,t+1}^{q}(x)
    \right]
    \\
    &=
    \sum_{t=1}^{L}\sum_{n=0}^{t-1}
    \omega_{n,t}^{q}(x_{<t})
    \Bigl[
        \nabla_\theta V_{n,t}^{q}(x)
        -
        A_{n,t}^{q}(x_{\leq t})\nabla_\theta V_{n,t+1}^{q}(x)
        -
        R_{n,t}^{q}(x_{\leq t})\nabla_\theta V_{t,t+1}^{q}(x)
    \Bigr]
    \\
    &=
    \sum_{t=1}^{L}\sum_{n=0}^{t-1}
    \omega_{n,t}^{q}(x_{<t})
    \left[
        \nabla_\theta c_{n,t}^{q}(x_{<t})
        +
        \nabla_\theta A_{n,t}^{q}(x_{\leq t})
        \left(
            V_{n,t+1}^{q}(x)-V_{t,t+1}^{q}(x)
        \right)
    \right].
\end{align*}

\subsection{Proof of Theorem \ref{thm:separation}}
\label{app:separation}

Let \(B\ge 3\) and let the vocabulary be
\(\mathcal V=\{0,1,a\}\).
We independently generate \(m\) subsequences
\(W_1,\ldots,W_m\), where
\[
    W_i =
    \begin{cases}
        0\underbrace{aa\cdots a}_{B-2 \text{ times}}0a, & \text{with probability } 2/3,\\
        1\underbrace{aa\cdots a}_{B-2 \text{ times}}1,  & \text{with probability } 1/3. 
    \end{cases}
\]
We refer to these as type-\(0\) and type-\(1\) subsequences, respectively.
Let the target distribution be 
\(X=W_1W_2\cdots W_m\eos\), so that
$     \E[L]=\left(B+\frac23\right)m $. 
The integer \(m\) will be chosen at the end of the proof.

We consider first-order drafters of the form
\[
    q_{n,t}\left(
        \cdot\mid x_{1:n};x_{n+1:t-1}
    \right)
    =
    q_{n,t}\left(
        \cdot\mid x_{1:n};x_{t-1}
    \right).
\]

\paragraph{The block-local optimum.}
Consider a round that starts exactly at the beginning of a
subsequence, following a committed prefix \(x_{1:n}\).
The first \(B-1\) draft positions can match the target
distribution exactly:
\[
    \widehat q_{n,n+1}
    =
    \frac23\delta_0+\frac13\delta_1,
    \qquad
    \widehat q_{n,n+j}
    =
    \delta_a,
    \quad 2\le j\le B-1.
\]
Here and below, these equalities are understood on
target-supported within-block prefixes.

At the final draft position \(n+B\), the target repeats the
first bit of the subsequence.
Since \(B\ge3\), the immediately preceding token is \(a\)
for both types.
A first-order drafter must therefore use the same distribution
at this position for both types.
To maximize the expected accepted length, $ \widehat q $ should maximize probablity of accepting the last token in the round, i.e, 
\begin{align*}
    \widehat q_{n, n+B} = \delta_0. 
\end{align*} 

If \(W_i=0\underbrace{aa\cdots a}_{B-2 \text{ times}}0a\), all \(B\) draft tokens are accepted,
and the target bonus token commits the final \(a\).
If \(W_i=1\underbrace{aa\cdots a}_{B-2 \text{ times}}1\), the proposal is rejected at the final
draft position, and residual sampling commits the final \(1\).
In either case, the round commits exactly the entire subsequence,
and the next round starts at the beginning of \(W_{i+1}\).
After the \(m\) subsequences have been generated, one additional
round generates \(\eos\).
Therefore, every block-local optimal drafter \(\widehat q\)
satisfies
\[
    \E[\cT^{\widehat q}]=m+1.
\]

\paragraph{A continuation-aware drafter.}
We next construct another first-order drafter \(q^*\) that
intentionally sacrifices accepted length in some rounds to
obtain more favorable future round starts.

Whenever a round starts at the beginning of a subsequence,
let \(q^*\) match the target distribution in the first
\(B-1\) positions:
\[
    q^*_{n,n+1}
    =
    \frac23\delta_0+\frac13\delta_1,
    \qquad
    q^*_{n,n+j}
    =
    \delta_a,
    \quad 2\le j\le B-1,
\]
but choose
\[
    q^*_{n,n+B}=\delta_1.
\]
If \(W_i=0\underbrace{aa\cdots a}_{B-2 \text{ times}}0a\), the final proposal is rejected.
The round commits the corrected \(0\) and advances by \(B\)
tokens, leaving the final \(a\) for the next round.
If \(W_i=1\underbrace{aa\cdots a}_{B-2 \text{ times}}1\), all \(B\) draft tokens are accepted.
The bonus token then commits the first token of the next
subsequence, or \(\eos\) if \(i=m\).

Whenever a round starts inside a subsequence, let \(q^*\)
match the target conditionals at every draft position.
This is possible within the first-order drafter class:
the bit defining the current subsequence is already contained
in the committed prefix.
Any fresh bit from the next subsequence appears at draft
position \(j\ge2\), and its repeated occurrence is at position
\(j+B-1\ge B+1\), outside the draft block.
Before that repeated occurrence, the intervening tokens are
all \(a\).
Thus no draft position needs access to an earlier within-block
bit beyond the immediately preceding token.
Every such round therefore advances by \(B+1\) tokens,
unless generation terminates earlier at \(\eos\).

To count the decoding rounds, extend \(W_1,W_2,\ldots\)
to an infinite independent sequence for this calculation.
Group consecutive subsequences between successive rounds
that start at the beginning of a subsequence.

If a group starts with a type-\(0\) subsequence, its first round
leaves the final \(a\) uncommitted.
Each subsequent round either leaves the final \(a\) of another
type-\(0\) subsequence or commits an entire type-\(1\)
subsequence and returns to a subsequence boundary.
The group therefore ends at the first type-\(1\) subsequence
after its initial type-\(0\) subsequence.
Its expected number of subsequences is \(1+3=4\), and it
uses exactly one round per subsequence.

If a group starts with a type-\(1\) subsequence, every round
advances by \(B+1\) tokens until a subsequence boundary is
reached again.
For any initial segment of this group, its total length is
\(B+1\) times its number of subsequences, minus its number
of type-\(1\) subsequences.
Consequently, the first return to a subsequence boundary
occurs upon completing the \((B+1)\)-st type-\(1\)
subsequence, counting the initial one.
The expected number of subsequences in this group is
\(1+3B\).
If the group contains \(\ell\) subsequences, its total length
is \((\ell-1)(B+1)\), so it uses exactly \(\ell-1\) rounds
and saves one round.

Let \(\ell_j\) denote the number of subsequences in group \(j\).
The groups are independent and identically distributed,
with
\[
    \E[\ell_j]
    =
    \frac23\cdot4+\frac13(1+3B)
    =
    B+3.
\]
Let
\[
    N
    :=
    \min\left\{
        k\ge1:\sum_{j=1}^{k}\ell_j\ge m
    \right\},
\]
and let \(H\) be the number of these \(N\) groups that
start with a type-\(1\) subsequence.
Since \(\{N\ge j\}\) depends only on the preceding groups
and \(N\le m\), summing expectations gives
\begin{align*}
    m
    &\le
    \E\!\left[\sum_{j=1}^{N}\ell_j\right]
    =
    \sum_{j=1}^{m}
    \E\!\left[
        \mathbf 1_{\{N\ge j\}}\ell_j
    \right]
    =
    (B+3)\E[N],
    \\
    \E[H]
    &=
    \sum_{j=1}^{m}
    \Pr\!\left(
        N\ge j,\,
        \text{group \(j\) starts with type \(1\)}
    \right)
    =
    \frac13\E[N].
\end{align*}

All groups except possibly the last are fully generated
before termination, and each completed type-\(1\) group
saves one round.
For the last group, suppose that \(r\) of its subsequences
remain before \(\eos\).
These subsequences contain at most \(r(B+1)\) tokens,
and only the first round of the group can lose one token
to rejection.
Thus generating these subsequences together with \(\eos\)
requires at most
\[
    \left\lceil
        \frac{r(B+1)+2}{B+1}
    \right\rceil
    =
    r+1
\]
rounds.
At most one of the \(H\) type-\(1\) groups is the last group,
so
\[
    \cT^{q^*}\le m+2-H.
\]
Taking expectations yields
\begin{align*}
    \E[\cT^{q^*}]
    &\le
    m+2-\E[H]
    =
    m+2-\frac13\E[N]
    \\
    &\le
    \left(1-\frac{1}{3(B+3)}\right)m+2.
\end{align*}

Finally, choose any integer
\(m>(3B+9)(3B+11)\).
Since both drafters preserve the same target distribution,
\begin{align*}
    \frac{\MAL(q^*)}{\MAL(\widehat q)}
    &=
    \frac{\E[\cT^{\widehat q}]}{\E[\cT^{q^*}]}
    \\
    &\ge
    \frac{m+1}
    {\left(1-\frac{1}{3(B+3)}\right)m+2}
    \\
    &>
    1+\frac{1}{3B+9}.
\end{align*}
This proves the theorem.

%% file: appendix.tex
\section{Additional Details for Expected Decoding Rounds}
\label{app:edr-details}

\subsection{Draft EOS Probability Does Not Reduce EDR}
\label{app:edr-eos}

For the EDR objective, a correctly drafted terminal $\eos$ does not save a
target-model verification round.  Conditional on reaching the terminal
position, accepting a drafted $\eos$ terminates in the current verification
pass.  If the draft instead misses $\eos$, the correction branch can generate
$\eos$ in that same pass and terminates before another round starts.  The same
holds when the terminal token is produced as the target bonus token after
block exhaustion.  Accordingly, the EDR sum runs only over ordinary-token
positions $t\le L$.

This is also visible in the dense local cost $ c_{n, t} $ .  Its numerator is
\[
    \sum_{v\in\cV}
    \pos{
        p_t(v\mid x_{<t})
        -q_{n,t}(v\mid x_{1:n};x_{n+1:t-1})
    },
\]
which excludes $\eos$. 
Thus matching target EOS mass is not directly rewarded by EDR. 

\subsection{Fixed-Size Anchor Sampling}
\label{app:importance-sampling}
In this section, we explain how to use importance sampling to select a fixed number of anchors in EDR training. 
We apply anchor sampling separately to each target rollout in
Algorithm~\ref{alg:edr}.
For a rollout \(x=(x_{1:L},\eos)\), define
\[
    \mathcal N_+(x)
    :=
    \left\{
        0\le n<L:
        \omega_{n,n+1}^{q_\theta}(x_{1:n})>0
    \right\}.
\]
An anchor outside \(\mathcal N_+(x)\) contributes zero to
\eqref{eq:edr-gradient}, since its occupancy
\(\omega_{n,t}^{q_\theta}(x_{<t})\) is zero for every \(t>n\).

Let \(\rho\) be the maximum number of training anchors per rollout,
set to \(512\) in our experiments.
If \(\rho\ge|\mathcal N_+(x)|\), select
\(\mathcal I=\mathcal N_+(x)\) and set \(\pi_n(x)=1\) for every
\(n\in\mathcal N_+(x)\).
Otherwise, choose \(\lambda(x)>0\) such that
\begin{equation}
    \sum_{n\in\mathcal N_+(x)}
    \min\left\{
        1,\lambda(x)\omega_{n,n+1}^{q_\theta}(x_{1:n})
    \right\}
    =\rho,
    \label{eq:anchor-lambda}
\end{equation}
and set
\begin{equation}
    \pi_n(x)
    :=
    \min\left\{
        1,\lambda(x)\omega_{n,n+1}^{q_\theta}(x_{1:n})
    \right\}.
    \label{eq:anchor-inclusion-prob}
\end{equation}
We draw a set \(\mathcal I\subseteq\mathcal N_+(x)\) of exactly
\(\rho\) anchors by random-start systematic sampling, with
\[
    \Pr(n\in\mathcal I\mid x)=\pi_n(x).
\]

For rollout \(x^{(i)}\), denote the sampled anchor set by
\(\mathcal I^{(i)}\) and write
\(\pi_n^{(i)}:=\pi_n(x^{(i)})\).
Using the same notation as Algorithm~\ref{alg:edr}, replace its
loss by
\begin{equation}
    \ell(\theta)
    =
    \frac1M\sum_{i=1}^{M}
    \sum_{n\in\mathcal I^{(i)}}\sum_{t=n+1}^{L_i}
    \operatorname{sg}\!\left(
        \frac{\omega^{(i)}_{n,t}}{\pi_n^{(i)}}
    \right)
    \left[
        c^{(i)}_{n,t}
        +
        A^{(i)}_{n,t}
        \operatorname{sg}\!\left(
            V^{(i)}_{n,t+1}-V^{(i)}_{t,t+1}
        \right)
    \right].
    \label{eq:anchor-sampled-loss}
\end{equation}
The occupancies and value functions are still computed by the
full recursions in Algorithm~\ref{alg:edr}, without gradients;
sampling restricts only which anchor contributions are differentiated.
The sampled anchor sets and inclusion probabilities are also
held fixed during differentiation.

Conditioned on the target rollouts, the inclusion probabilities
cancel the inverse-probability weights. Hence,
\begin{align}
    &\E_{\{\mathcal I^{(i)}\}_{i=1}^{M}}\!\left[
        \nabla_\theta\ell(\theta)
        \,\middle|\,
        \{x^{(i)}\}_{i=1}^{M}
    \right]
    \notag\\
    &\qquad=
    \frac1M\sum_{i=1}^{M}
    \sum_{n\in\mathcal N_+(x^{(i)})}
    \sum_{t=n+1}^{L_i}
    \omega^{(i)}_{n,t}
    \left[
        \nabla_\theta c^{(i)}_{n,t}
        +
        \nabla_\theta A^{(i)}_{n,t}
        \left(
            V^{(i)}_{n,t+1}-V^{(i)}_{t,t+1}
        \right)
    \right]
    \notag\\
    &\qquad=
    \frac1M\sum_{i=1}^{M}
    \nabla_\theta
    \widehat{\mathcal L}_{\mathrm{EDR}}
        (q_\theta;x^{(i)}),
    \label{eq:ht-gradient-unbiased}
\end{align}
where the last equality follows from \eqref{eq:edr-gradient}.
Taking the expectation over the target rollouts and applying
\eqref{eq:edr-gradient-unbiased} gives
\[
    \E\!\left[\nabla_\theta\ell(\theta)\right]
    =
    \nabla_\theta\mathcal L_{\mathrm{EDR}}(q_\theta).
\]
Thus anchor sampling preserves preserves the unbiasedness of the stochastic gradient.

%% file: exp_details.tex
\section{Additional Details for Experiments}
\label{app:exp-details}


For all training settings and methods, we use the hyperparameters summarized in Table~\ref{tab:training-hyperparameters}.

\begin{table}[h]
    \centering
    \caption{Training hyperparameters used in all experiments.}
    \label{tab:training-hyperparameters}
    \begin{tabular}{ll}
        \toprule
        \textbf{Hyperparameter} & \textbf{Value} \\
        \midrule
        Learning rate        & $3\times10^{-5}$ \\
        Optimizer            & AdamW \\
        AdamW $\beta_1$      & $0.9$ \\
        AdamW $\beta_2$      & $0.999$ \\
        Learning-rate schedule & $4\%$ linear warmup of the first epoch, then constant \\
        Gradient clipping    & Global norm $\leq 1$ \\
        Global batch size    & $96$ \\
        \bottomrule
    \end{tabular}
\end{table}